# Using Twitter to predict football outcomes

Stylianos Kampakis, University College London, stylianos.kampakis@gmail.com

Andreas Adamides, University College London

**Abstract**

Twitter has been proven to be a notable source for predictive modeling on various domains such as the stock market, the dissemination of diseases or sports outcomes. However, such a study has not been conducted in football (soccer) so far. The purpose of this research was to study whether data mined from Twitter can be used for this purpose. We built a set of predictive models for the outcome of football games of the English Premier League for a 3 month period based on tweets and we studied whether these models can overcome predictive models which use only historical data and simple football statistics. Moreover, combined models are constructed using both Twitter and historical data. The final results indicate that data mined from Twitter can indeed be a useful source for predicting games in the Premier League. The final Twitter-based model performs significantly better than chance when measured by Cohen's kappa and is comparable to the model that uses simple statistics and historical data. Combining both models raises the performance higher than it was achieved by each individual model. Thereby, this study provides evidence that Twitter derived features can indeed provide useful information for the prediction of football (soccer) outcomes.

## Introduction

The first launch of Twitter took place on July 2006. Over the period of the last eight years, Twitter has become a big player in the social networking industry with an extremely large user-base, consisting of several millions of users. In particular, the number of Twitter accounts up to mid 2013 reached the number of 500 million user accounts. However the estimated number of active users was around 200 million, while the daily number of tweets, climbed up to 500 million[1].

Twitter has been used for predicting or explaining a variety of other real-world events, such as the outcome of elections (Tumasjan, Sprenger, Sandner, & Welpe, 2010), the stock market (Bollen, Mao, & Zeng, 2011), box-office revenues of movies in advance of their release (Asur & Huberman, 2010) and the spread of diseases (Paul & Dredze, 2011). These are all strong evidence that Twitter can be a source of meaningful and useful information which can be exploited by using statistical methods.

---

[1] www.telegraph.co.uk/technology/twitter/9945505/Twitter-in-numbers.html

The Premier League is a professional football league and it consists of twenty teams. Official metrics for the 2010-2011 football season reported an audience of 4.7 billion people in 212 countries/territories and with 80 different broadcasters[2].

A regular season runs every year from August until May of the next year, with each team playing thirty-eight matches in total. The teams that take part in this competition have matches at least once a week throughout the season and more importantly all of these matches are broadcast on a nation-wide basis and not only. This opens the opportunity for many people to comment, express their opinions or to engage in further conversations on social media websites, such as Twitter.

Twitter can be proven to be a valuable source of information, simply because it can provide, in some cases, more informative data than the ones found in other statistical or historical sources, or at some cases data that do not exist at all. A few examples are players' injuries, sacked coaches, and the average sentiment amongst the fan base of each team. This study attempts to identify to what extent information mined from Twitter can be useful for predicting football outcomes.

This is done through three different models. First, we assess the performance of model that is based solely on Twitter features. Then, we build a model that is based on historical and other statistics about the football teams and we compare this model with the Twitter model. Finally, we combine the both approaches in order to understand whether Twitter can provide any additional information on top of the information that is already available by historical statistics.

We find that the Twitter-extracted features, in particular n-grams used in a bag-of-words, can indeed be a reliable source for predicting football outcomes. The final Twitter model achieved an average accuracy score of around 65%, which is comparable to the model that used simple historical statistics and had an average accuracy of around 59%. However, even higher accuracy is achieved if features from both datasets are combined, with the final accuracy going up to 75%. The results demonstrate that information mined from Twitter can be useful for predicting football outcomes, a fact that certainly opens opportunities for further research.

## Related work

There is a great deal of work that has been done in predicting sports outcomes. Here we present some relatively research on the field. Sinha et al. (Sinha, Dyer, Gimpel, & Smith, 2013) used n-grams derived from Twitter in order to predict outcomes of the National Football League (NFL) and compared the performance of this model to a model that uses simple statistics. They concluded that the Twitter model can match or exceed the performance of statistics-only model.

Another work (Lock & Nettleton, 2014) combined a number of situational variables for each team in NFL games (remaining yards, field position, current score etc.) into a random forest classifier. They achieved a good accuracy score when predicting outcomes, especially at the later stages of a game. Another interesting project in this context is *TwitterPaul* (UzZaman, Blanco, & Matthews, 2012), a system which retrieved filtered tweets about the 2010 Fifa World Cup with the aim of making predictions about the outcome of a match. The final model reached an accuracy of up to 88%.

---

[2] http://bleacherreport.com/articles/1948434-why-the-premier-league-is-the-most-powerful-league-in-the-world

On a direction of modelling the outcomes of football games based on historical data, Rue and Salvesen (Rue & Salvesen, 2000) used a Bayesian generalized linear model to predict outcomes with reasonably good results. Baio and Blangiardo (Baio & Blangiardo, 2010) used a Bayesian statistical model with data from the 1993-1994 season and then applied that model to predict outcomes for each team (overall points, attack/defence goals) for the 2007-2008 season. The final model produced relatively accurate results.

## Data

The data consists of N games collected over the period 21st of March - 11th of May 2014. There were three datasets: the twitter dataset, the historical and simple statistics dataset (abbreviated to historical dataset from now on) and the combined dataset. The combined dataset is simply a merge of the first two. The twitter dataset and the historical dataset are described below.

### Twitter dataset

This dataset consisted of around 2 million fan's tweets about each of their favourite teams and was retrieved through Twitter's open Streaming API.

Using the Twitter streaming API the data gathering started on Friday the 21st of March and lasted until the 11 of May, which was actually the day of the last match in the league and also signalled the end of the football Premier League. We managed to get information for a range of eight to ten matches for each team, since every club did not have the same number of matches to play during this period. The exact final number of retrieved tweets for all twenty teams was 1,975,614.

We created a list of hashtags which are strongly associated with each of the twenty teams in Premier League. The creation of this list was for the most part based in several online resources which identify the official hashtags for the teams along with any nicknames that a team might have[3,4]. There were variations across teams; a snippet of our gathered list for five teams of the Premier League can be seen at figure 1.

| Team | Pre-Defined Hashtags | | | | | |
|---|---|---|---|---|---|---|
| Arsenal | #arsenal | #arsenalfc | #afc | #gunners | #thegunners | #arsenalfamily |
| | #gooners | #comeongunners | #comeongooners | #coyg | #comeonarsenal | #afcfamily |
| Aston Villa | #astonvilla | #astonvillafc | #avfc | #villans | #comeonastonvilla | #comeonvilla |
| | #avfcfamily | #comeonyouvillans | #villafamily | #govilla | #goastonvilla | #govillans |
| Everton | #everton | #evertonfc | #efc | #toffees | #efcfamily | #goeverton |
| | #evertonfamily | #schoolofscience | #comeoneverton | #coyb | #comeontoffees | #gotoffees |
| Manchester City | #manchestercity | #manchestercityfc | #mancity | #citizens | #mancityfc | #mcfc |
| | #gomcfc | #gomancity | #comeonmancity | #mcfcfamily | #comeoncitizens | #mancityfamily |
| Swansea | #swanseacity | #swanseacityfc | #swanseafc | #swans | #swansfc | #jackarmy |
| | #comeonswansea | #twitterjacks | #comeonyoujacks | #goswansea | #comeonyouswansea | #gojacks |

Figure 1. Examples of hashtags associated with some of the teams in the Premier League.

---

[3] www.soccer-corner.com/en-gb/football-clubs
[4] www.premierleague.com/en-gb/kids/clubs.html

We discarded tweets that contained hashtags for more than one team. The main reason for this was that we wanted to maintain focus on tweets that were basically comments on particular games from the perspective of one of the two teams. This would also result in a massive duplication across the tables in our database for each of the teams, therefore we filtered out those instances. Thus, if a tweet contained any number of hashtags corresponding to exactly one Premier League team, the tweet was assigned to that team only and was used for the data analysis. If we were going to include those, then there was the danger of including tweets that were not associated with only one team.

Moreover, during this process we discovered some other facts that needed to be addressed. For example we found that there is an obvious correlation between the nicknames of a few Premier League teams and other sports teams from around the world. It is a fact that the nickname of the team of Southampton F.C., Saints is identical to the nickname of the National Football League (NFL) team of the New Olreans Saints. Also, this is the case for the Premier League team of Tottenham Hotspurs, having the same nickname, Spurs, as the National Basketball Association (NBA) team of San Antonio Spurs. We tackled the above issues by simply filtering out those tweets. If we kept them in, our final results would not be representable with respect to each of these two teams.

We noticed that there is a significant skew in the total number of tweets amongst each of the twenty teams. Nevertheless, this is quite reasonable because a few teams out of all are much more famous, well-known, successful within the league and it is also known that they have a greater number of fans than other teams. The team with the most retrieved tweets for this period is Liverpool F.C. with 426,457 tweets, the top second is Manchester United with 416,767 tweets while the teams with the lowest number of retrieved tweets is Fulham with 15,530 and then Swansea City with 15,668 mentions. Figure 2 shows the bar charts of the tweets for each different club in the Premier League.

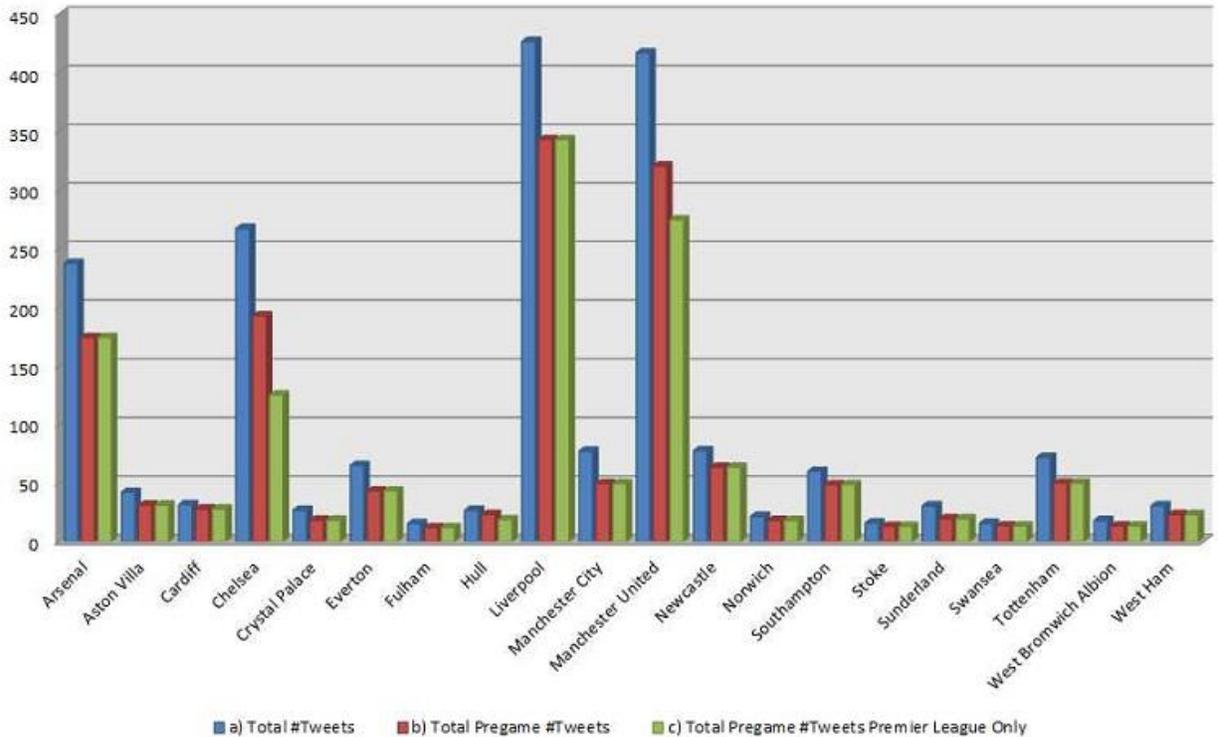

**Figure 2. Bar charts of the tweets for each team in the premier league**

The data were processed by using the TwitterNLP and Part-of-Speech Tagger published by the ARK social media research group in Carnegie Mellon. We only included words with the following tags: Adjective, Verb, Noun, Adverb, Interjection, Emoticons or Possesive. We then stemmed the words in the dataset by using the Java implementation of the Porter stemmer. Finally, we split the dataset in two versions, one with unigrams and one with bigrams.

### Historical dataset

We acquired a variety of historical data and simple statistics for the English Premier League. The statistics were collected over the exact same period as the Twitter dataset.

Table 1 presents the features that were used. Features 1-6 and $F_8$ represent averages that had to be updated before each game in the dataset. The rest of the statistics (total market value for example) remained unchanged.

Table 1. Table of features for the historical dataset

| Feature | Description |
|---|---|
| $F_1$ = Avg. Goals | Average number of scored goals. |
| $F_2$ = Avg. Corners | Average of earned corners. |
| $F_3$ = Avg. Shots On Target | Average number of shots on target. |
| $F_4$ = Avg. Fouls Committed | Average fouls committed per team. |
| $F_5$ = Avg. Yellow Cards | Average Bookings with yellow cards. |
| $F_6$ = Avg. Red Cards | Average Bookings with yellow cards. |
| $F_7$ = Market Value | Total Market Value of each team in millions. |
| $F_8$ = Avg. Squad Age | Average Age of players in the squad. |
| $F_9$ = No. International Players | Total number of players in the squad that play for their home country too |
| $F_{10}$ = EPL Wins | Total number of wins for the Premier League, otherwise 0 if a team did not ever win the EPL. |
| $F_{11}$ = EPL Runners-Up | Total number of second place finish, otherwise 0 if a team did not ever finish second. |
| $F_{12}$ = Double Winners | Total number of times a team won both the Premier League and FA Cup, otherwise 0 if a team did not win a double. |

## Methods

The purpose of the study was the prediction of the outcome of a game as being a win for the home team, a win for the away team or a draw. Each instance of the dataset represents a match. The instance is split into three parts:

- Home team features
- Away team features
- Response variable

So, the input vector consists of two parts. The home features $f_h$ and the away features $f_a$. The input for the classifier is the vector $f_{input} = [f_h, f_a]$.

Regarding the types of features, there were three variations, depending on the dataset:

- Bag-of-words (twitter dataset)
- Historical features
- Combined (bag-of-words plus historical features)

For the Twitter model, the home and away features correspond to a bag of unigrams or bigrams. For the historical model, the features correspond to the statistics for each club.

The bag-of-words for the Twitter was pre-processed through the use of chi-square and the mutual information index. We calculated the p-value of a chi-square test and the mutual information for each unigram and bigram and we ranked them according to importance. The machine learning models were then tested on different numbers of unigrams and bigrams, from 1 to 35. We ended up having a different bag-of-words for the home teams and a different bag-of-words for the away teams. So, for example,

Liverpool and Manchester United will have the same features in their bag-of-words when playing at home, but Liverpool (as well as Manchester United) will have different features in the bag-of-words when playing away.

The models used were Naïve Bayes, Random forests, Logistic regression and SVM. All the models were tested through leave-one-out cross-validation. Because random forests make use of a random seed that can affect the final result, we conducted 100 rounds of cross-validation for random forests, using a different random seed each time. Random forests were used with different numbers of trees from 10 up to 1000. The SVM was tested with a radial basis function kernel, a sigmoid kernel and a polynomial kernel and various parameter settings for each kernel.

The results were evaluated through the use of the accuracy and Cohen's kappa. For the random forests we used the out-of bag error as an estimate of the accuracy on the test set. Cohen's kappa is particularly useful for assessing the performance of a classifier against a classifier that is guessing at random. Values close to zero indicate that the classifier is no better than a random classifier and values close to 1 indicate perfect performance.

## Results

Table 2 below summarizes the best results for each dataset. The statistics provided are the mean, the standard deviation, the minimum and the maximum for the accuracy and the Cohen's kappa statistic.

**Table 2. Best classifier for each dataset.**

| Model | Best classifier | Mean Kappa and Accuracy | Min | Max |
|---|---|---|---|---|
| Twitter-only | Random forest | 0.25+/-0.093 | 0.047 | 0.389 |
|  |  | 65.6%+/-4.33% | 56.3% | 74.7% |
| Historical-only | Naïve Bayes | 0.239+/-0.075 | 0.146 | 0.315 |
|  |  | 58.9%+/-5.97% | 50.6% | 64.4% |
| Combined | Random forest | 0.28+/-0.065 | 0.144 | 0.418 |
|  |  | 69.6%+/-2.4% | 64.4% | 74.7% |

Regarding the twitter model, the random forest was the best classifier. Figure 3 below shows the performance of the random forest for different number of bigrams and unigrams, filtered by chi-square or information gain. The x-axis shows the number of bigrams or unigrams for each team. So, the value 10 on the x-axis means the total number of features was 20, since the model used a bag-of-words of size 10 for the home team and another bag-of-words of size 10 for the away team.

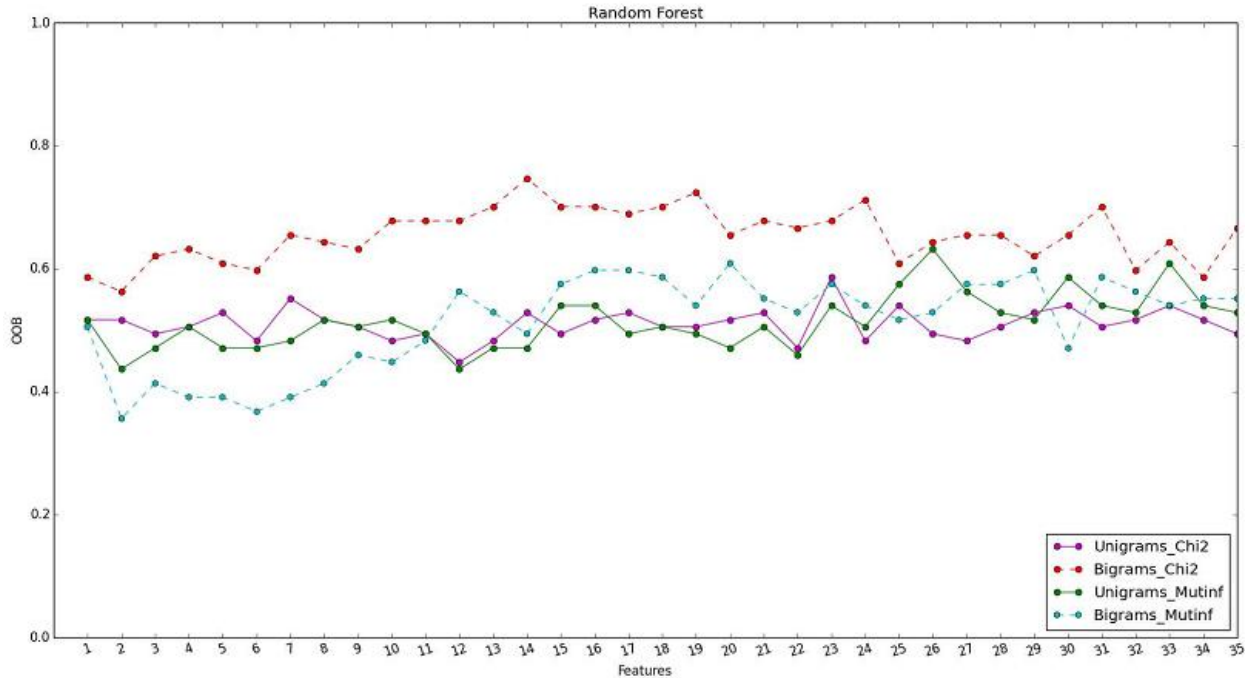

**Figure 3. Out-of-bag error for the random forest on the twitter dataset**

The figure clearly shows that the best results are achieved using bigrams derived by chi-square. It seems that the performance keeps rising until we reach about 11-15 features for each side, and then starts falling.

Regarding the historical dataset, the best classifier was Naïve Bayes. The accuracy of Naïve Bayes seems to be lower than the accuracy achieved by the random forest on the Twitter dataset, but the Kappa statistics are very close together. This might be an indication that the random forest for the Twitter dataset might predict the majority class quite often, leading to a high accuracy score at the expense of misclassifying quite a few examples.

For the combined model, random forest was the best classifier. Figure 4 shows the results of the random forest on the combined dataset for different numbers of bigrams and unigrams. Like we observed on the Twitter dataset, it seems that the optimal performance is achieved when around 11-15 bigrams are used. The additional historical features, however, improved both the accuracy and the kappa statistic of the classifier. The final kappa statistic was 0.28 in contrast with 0.25 for the Twitter-only model and 0.239 for the historical model. Similarly, the average accuracy improved to 69.6% from 65.6% for the Twitter-only model and 58.9% for the historical model.

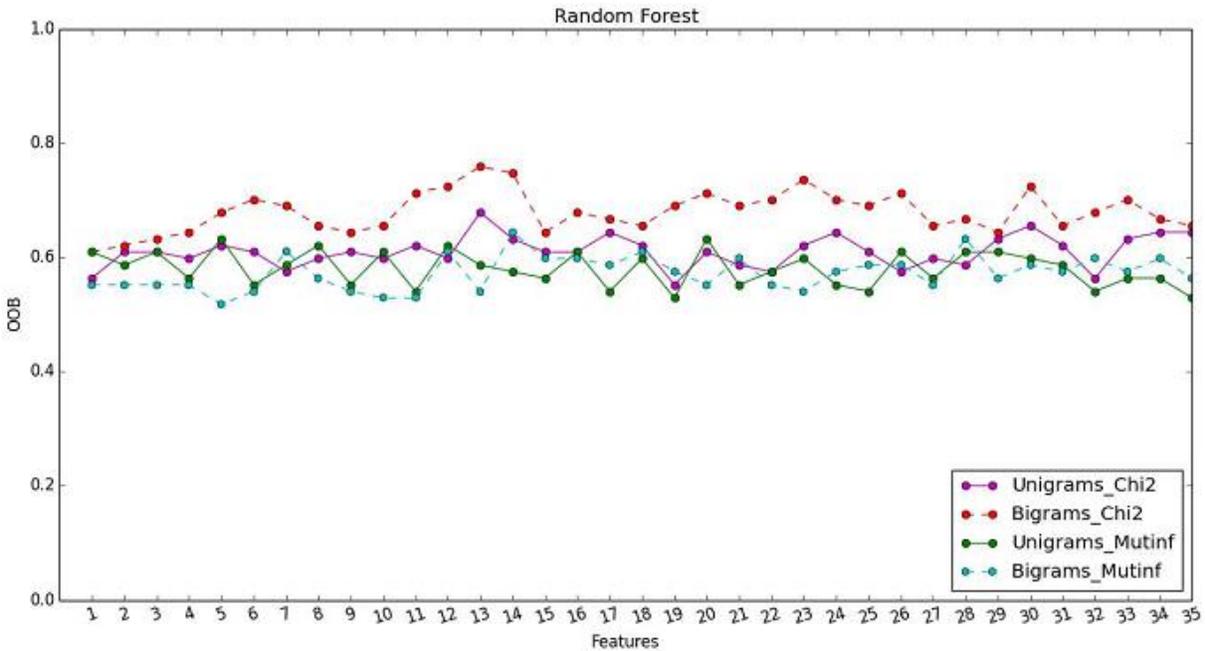

**Figure 4. Out-of-bag error for the random forest on the combined dataset**

## Discussion and conclusion

Something that becomes clear from the results is that Twitter contains enough information to be useful for predicting outcomes in the Premier League. Furthermore, it seems that information contained in Twitter is not available in simple statistics, since the combination of the two datasets improved the results beyond what could be achieved by each individual dataset.

In our results we can't neglect a number of limitations of this study. First and foremost, the experiment was not conducted on the whole English Premier League Season, but instead on a part of it[5]. It would be interesting to perform the same study but on a complete season. Another interesting approach would be to use models built on data from one season to predict the results of the next.

In a similar context, acquiring historical data proved to be extremely difficult and time-consuming. The fact that the results on the historical dataset are not as good as on the Twitter dataset, could be a side-effect of the sources or features that were used. Therefore, it is possible to consider the current study a complete comparison between the use of historical statistics and Twitter-mined features for predictive modeling. Nevertheless, this fact does not limit the conclusion of the study which is that Twitter contains information which is useful for predictive modeling.

It was observed that bigrams perform better than unigrams, which indicates that the information contained in Twitter is more complex than simple unigrams can provide. It would be interesting to see whether higher order n-grams (trigrams, etc.) could perform even better.

---

[5] An application was directly made to Twitter where they had data grants/openings for academic purposes, but with receiving a negative response.

An interesting question that can be posed, based on the results of this study, is what exactly was the nature of the information that helped in the predictions. In the introduction we speculated that there can be different useful sources of information: the emotion of the fans, discussions about the outcome or news stories, such as injuries, which are not included in a statistical model. Future research could clarify which sources are used in the prediction and to what extent.